\documentclass[conference]{IEEEtran}
\IEEEoverridecommandlockouts
\usepackage{cite}
\usepackage{amsmath,amssymb,amsfonts}
\usepackage{algorithmic}
\usepackage{graphicx}
\usepackage{textcomp}
\usepackage{xcolor}
\usepackage{cite}
\usepackage{amsmath,amssymb,amsfonts}
\usepackage{algorithmic}
\usepackage{graphicx}
\usepackage{textcomp}
\usepackage{xcolor}
\usepackage{booktabs}
\usepackage{multirow}
\usepackage{tabularx}
\usepackage{color}
\usepackage{colortbl}
\usepackage{tcolorbox}
\usepackage{url}
\usepackage{utfsym}
\usepackage{caption}
\usepackage{pifont}
\newcommand{\cmark}{\ding{51}}  % Check mark
\newcommand{\xmark}{\ding{55}}  % Cross mark
\def\BibTeX{{\rm B\kern-.05em{\sc i\kern-.025em b}\kern-.08em
    T\kern-.1667em\lower.7ex\hbox{E}\kern-.125emX}}
\begin{document}

\title{DialogueAgents: A Hybrid Agent-Based Speech Synthesis Framework for Multi-Party Dialogue}

\author{\IEEEauthorblockN{Xiang Li$^{1*}$\thanks{\(^*\) Equal Contribution, \(^\dag\) Corresponding Author. }, Duyi Pan$^{2*}$, Hongru Xiao$^{2}$, Jiale Han$^{3}$, Jing Tang$^{2}$, Jiabao Ma$^{1}$, Wei Wang$^{2,3\dag}$, Bo Cheng$^{1}$}
\IEEEauthorblockA{
\textit{$^1$State Key Laboratory of Networking and Switching Technology, Beijing University of Posts and Telecommunications} \\
\textit{$^2$Department of Computer Science and Engineering, Hong Kong University of Science and Technology (Guangzhou)} \\
\textit{$^3$Hong Kong University of Science and Technology} \\
\{lixiang2022,jiabao.m,chengbo\}@bupt.edu.cn, \{jialehan,weiwcs\}@ust.hk, 
\{duyipan,hongruxiao,locusjingt\}@hkust-gz.edu.cn}
}

\maketitle

\begin{abstract}
Speech synthesis is crucial for human-computer interaction, enabling natural and intuitive communication. However, existing datasets involve high construction costs due to manual annotation and suffer from limited character diversity, contextual scenarios, and emotional expressiveness. To address these issues, we propose DialogueAgents, a novel hybrid agent-based speech synthesis framework, which integrates three specialized agents—a script writer, a speech synthesizer, and a dialogue critic—to collaboratively generate dialogues. Grounded in a diverse character pool, the framework iteratively refines dialogue scripts and synthesizes speech based on speech review, boosting emotional expressiveness and paralinguistic features of the synthesized dialogues. Using DialogueAgent, we contribute MultiTalk, a bilingual, multi-party, multi-turn speech dialogue dataset covering diverse topics. Extensive experiments demonstrate the effectiveness of our framework and the high quality of the MultiTalk dataset. We release the dataset and code\footnote{\url{https://github.com/uirlx/DialogueAgents}} to facilitate future research on advanced speech synthesis models and customized data generation.

\end{abstract}


\section{Introduction}
\label{sec:intro}

Speech synthesis plays an important role in enhancing human-computer interaction by converting natural language into human-like audio. However, the complexity of real-life conversational scenarios presents significant challenges for speech synthesis, particularly in situations involving multiple participants and multi-turn dialogues\cite{DBLP:conf/interspeech/SaitoTITS23}. Effectively maintaining coherence across different dialogue turns and accurately expressing the emotions of multiple participants in the context are critical issues that need to be addressed in the field.

%However, existing speech synthesis datasets primarily focus on synthesizing individual utterances without any dialogue context. Although a few dialogue datasets\cite{DBLP:conf/icassp/LeePK23} have been released recently, most of them are created by professional voice actors transcribing meticulously crafted dialogue scripts, leading to relatively high construction costs. Alternatively, these datasets often consist of dialogues with limited participants and turns, and fail to adequately capture emotional expression, as presented in Table~\ref{table:data}. Therefore, there is an urgent need for a more natural and emotionally expressive multi-party multi-turn dialogue dataset with a diverse range of roles.

%Although a few dialogue datasets have been released recently, most of them \cite{DBLP:conf/icassp/LeePK23} are created by professional voice actors transcribing meticulously crafted dialogue scripts, leading to relatively high construction costs. Moreover, these datasets \cite{DBLP:conf/lrec/CieriMW04} consist of dialogues with limited participants and turns, failing to adequately capture emotional expression and lacking role diversity. KE-Omni \cite{zhao2024advancing} is an attempt to automate dataset synthesis through sequential cooperation of textual and speech agents; however, may lead to accumulated errors and less effectiveness. Therefore, there is an urgent need for a more natural and emotionally expressive multi-party multi-turn dialogue dataset with a diverse range of roles.

However, existing speech synthesis datasets \cite{DBLP:conf/ococosda/VeauxYK13, ma2024wenetspeech4tts} primarily focus on synthesizing individual utterances without incorporating dialogue context. While a few dialogue datasets have been released recently, most of them \cite{DBLP:conf/icassp/LeePK23} are created by manual transcription, leading to relatively high construction costs. Moreover, these datasets \cite{DBLP:conf/lrec/CieriMW04} consist of dialogues with limited participants and turns, failing to adequately capture emotional expression and lacking role diversity. KE-Omni \cite{zhao2024advancing} is an attempt to automate dataset synthesis through sequential cooperation of textual and speech agents; however, this sequential approach may lead to accumulated errors and reduced effectiveness. Therefore, there is an urgent need for better automatic methods to create multi-party and multi-turn speech dialogue datasets.

%Current speech synthesis tasks \cite{DBLP:conf/iclr/0006H0QZZL21} primarily focus on synthesizing individual utterances without considering the impact of context on speech, leading to subpar performance in dialogue scenarios. In addition, we observe that existing datasets mainly consist of single-text synthesis without any dialogue context. Although a few dialogue datasets \cite{DBLP:conf/icassp/LeePK23} are released recently, they remain limited in role diversity and fail to adequately capture paralinguistic information such as pauses, laughter, and emotional expression. Therefore, there is an urgent need for a more natural and emotionally expressive multi-party multi-turn dialogue dataset with a diverse range of roles.

\begin{table}[t]
\caption{Comparison of MultiTalk with existing speech synthesis datasets. Party represents the involvement of multiple parties (more than two participants); Turns indicates the inclusion of multiple turns (exchanges) in the dialogues.}
\label{table:data}
\centering
\renewcommand{\arraystretch}{0.88}
\renewcommand{\tabcolsep}{1.2pt}
\resizebox{0.45\textwidth}{!}{
\begin{tabular}{lcccc}
\toprule
\textbf{Dataset} & \textbf{Dialogue} & \textbf{Party} &  \textbf{Turns}  & \textbf{Creation} \\
\midrule
VCTK \cite{DBLP:conf/ococosda/VeauxYK13} & \xmark & \xmark & \xmark & Manual transcription \\
WS4TTS\cite{ma2024wenetspeech4tts} & \xmark & \xmark & \xmark  & Manual transcription \\
Fisher \cite{DBLP:conf/lrec/CieriMW04} & \cmark & \xmark & \cmark  & Telephone recordings \\
DailyTalk \cite{DBLP:conf/icassp/LeePK23} & \cmark & \xmark & \cmark & Manual transcription \\
KE-Omni \cite{zhao2024advancing} & \cmark & \xmark & \xmark  & Sequential agents synthesis \\
\textbf{MultiTalk} & \cmark & \cmark & \cmark  & Iterative agents synthesis \\
\bottomrule
\end{tabular}}
\end{table}

To this end, we propose DialogueAgents, a hybrid agent-based speech synthesis framework, capable of automatically generating speech dialogues with natural turn-taking, contextual emotional expression, and diverse character involvement. Specifically, our framework employs three intelligent agents to generate dialogue scripts, corresponding speech, and reviews of the synthesized speech. Given a pool of distinctive characters, the framework first utilizes a Script Writer Agent to select participants from the character pool and generate the dialogue script, followed by a Speech Synthesizer Agent that converts the script into spoken dialogue. Moreover, a Dialogue critic Agent is introduced to review the synthesized speech from multiple perspectives and provide optimization feedback. Upon receiving this feedback, the Script Writer refines the script by incorporating paralinguistic and emotional markers. This entire process undergoes multiple iterations, leveraging the collaborative capabilities of the hybrid agents to continuously refine the script and dialogue speech.

Based on our proposed DialogueAgents, we generate a bilingual (Chinese and English) multi-party multi-turn speech dialogue dataset, namely MultiTalk, as presented in Table~\ref{table:data}. Additionally, we design two dialogue-level speech evaluation metrics, Emotional MOS (EMOS) and Turn-taking MOS (TMOS), to reflect the emotional expressiveness throughout the dialogue and the naturalness of turn-taking, respectively. Extensive experiments demonstrate the effectiveness of our speech dialogue synthesis framework and the quality of the generated dataset.

Our main contributions are as follows:
\begin{itemize}
  \item We propose a novel hybrid agent-based speech synthesis framework  for multi-party dialogue systems, leveraging the reflection and refinement among different agents to iteratively boost the synthesis quality.
  \item We contribute a bilingual speech dialogue dataset with diverse participants and rich paralinguistic information for multi-party multi-turn conversational scenarios.
  \item Qualitative and quantitative experiments validate the effectiveness of the framework and the integrity of the dataset.
\end{itemize}

\section{Related Work}
\begin{figure*}[t]  % 通栏的带*号
    \centering
    \includegraphics[width=0.63\textwidth]{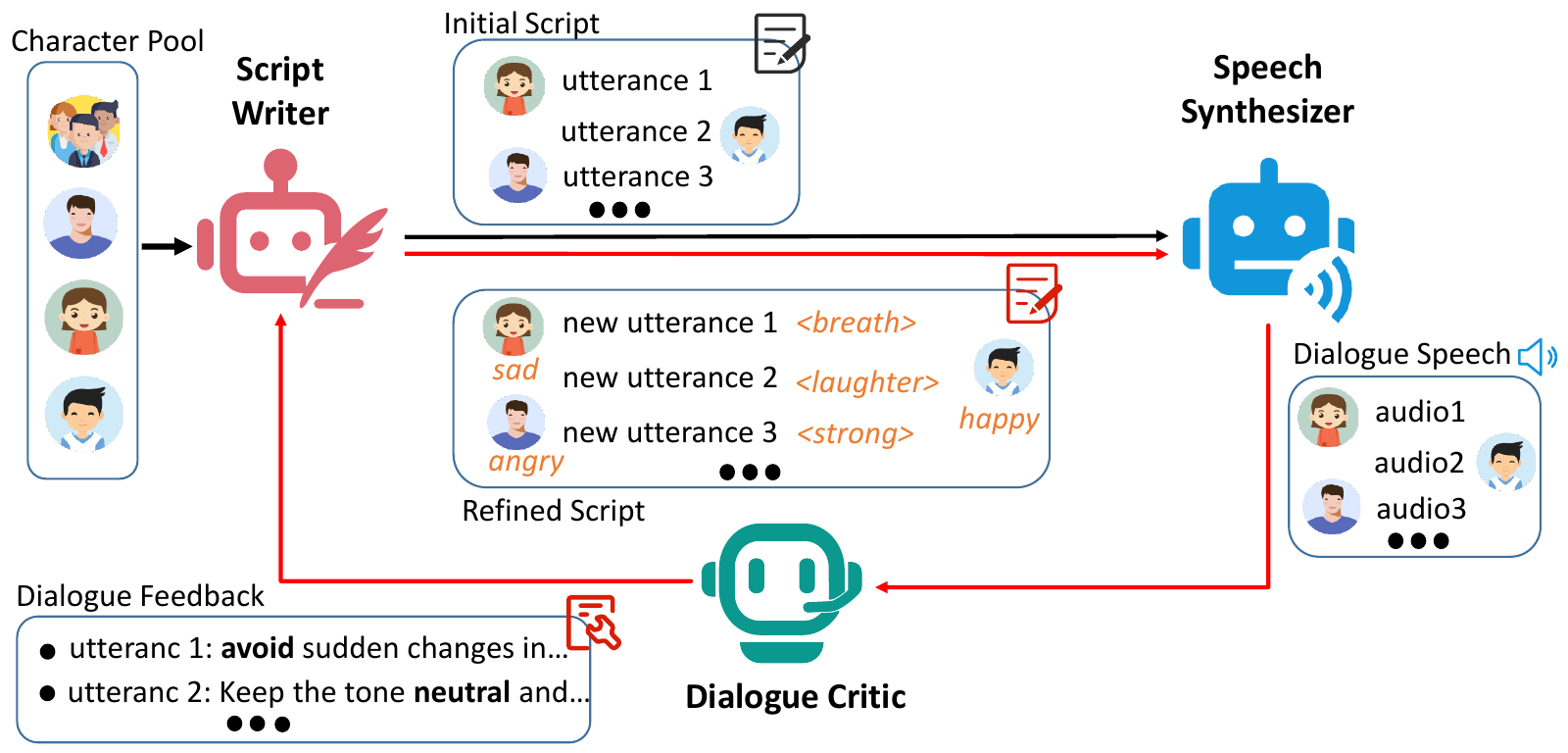}
    % \vspace{-0.5cm}
    \caption{The overall framework of DialogueAgents.}
    \label{fig:framework}
    % \vspace{-0.3cm}
\end{figure*}

\subsection{Dialogue Speech Synthesis Models}
Recent advancements in speech synthesis model have primarily focused on integrating a few seconds of a speaker's audio as a prompt to guide the synthesis process\cite{DBLP:conf/iclr/ShenJ0LL00Z024}. This paradigm has demonstrated excellent performance in zero-shot speech synthesis. However, most of these methods concentrate on high-fidelity cloning of individual utterances, with limited exploration and extension to dialogue speech synthesis, particularly in multi-speaker interaction scenarios. %dGSLM \cite{DBLP:journals/tacl/NguyenKCAHETASM23} represents the first ``text-free'' model designed to generate natural dialogue audio samples. Utilizing a dual-tower Transformer architecture with cross-attention mechanisms, it is trained solely on 2,000 hours of dual-channel raw dialogue audio from the Fisher dataset, without requiring any text or labels. 
ChatTTS, a model specifically designed for dialogue synthesis, has surpassed most open-source speech synthesis models in terms of fluency and naturalness. However, the training data is not publicly available for researchers, and the models remain somewhat unstable when generating long dialogues and multi-speaker interactions. Benefiting from the enhancement provided by large language models, multi-modal models like Qwen2-Audio have shown promise in utilizing paralinguistic information, but they face challenges in supporting real-time, duplex speech interaction. On the other hand, models like GPT-4o have demonstrated strong low-latency dialogue interaction abilities. Despite these advancements, both systems are still constrained by the scarcity of high-quality dialogue training data, which limits their ability to provide seamless interaction and a naturally fluent user experience.

\subsection{Dialogue Speech Synthesis Dataset}
In previous research, the Fisher \cite{DBLP:conf/lrec/CieriMW04}  Dataset was introduced, collected from over 16,000 English telephone conversations, each averaging ten minutes in length and covering a range of topics. However, the audio is limited to two separate channels. DailyDialog  \cite{DBLP:conf/ijcnlp/LiSSLCN17}, a high-quality multi-turn dialogue dataset, is human-written and relatively noise-free. The dialogues in the dataset reflect typical patterns of daily communication and cover various topics related to everyday life. However, it lacks corresponding audio resources. DailyTalk \cite{DBLP:conf/icassp/LeePK23} sampled and adapted 2,541 dialogues from the DailyDialog dataset, which were recorded by two English-fluent speakers. The actors followed emotion labels and were instructed to naturally incorporate filler words (e.g., "uh," "um") in approximately half of the dialogues, thereby enhancing the TTS model's ability to generate such expressions. However, the dialogues are limited to interactions between two speakers and are conducted exclusively in English. The recently proposed AnyInstruct \cite{DBLP:conf/acl/ZhanDYZZLZYZL0F24} dataset consists of 108,000 spoken dialogues generated using the Azure Text-to-Speech API, featuring 39 distinct timbres. However, the dataset is primarily focused on human-machine interaction and is exclusively in English, which limits its applicability for research and applications in multi-speaker or multilingual dialogue contexts

\subsection{Multi-Agent Collaboration}
Previous research on multi-agent systems has primarily focused on enhancing reasoning by facilitating collaboration among multiple models \cite{DBLP:conf/coling/GuHWL025}. Tencent’s MAD framework \cite{DBLP:conf/emnlp/Liang0JW00Y0T24} addresses the Degeneration-of-Thought problem by allowing agents to argue and be judged on their responses. ChatEval \cite{DBLP:conf/iclr/ChanCSYXZF024} evaluates model responses autonomously, while ReConcile \cite{DBLP:conf/acl/ChenSB24} uses a roundtable setup for collaborative reasoning with weighted voting. These efforts demonstrate that multi-agent collaboration can boost performance by simulating human-like cognitive synergy for complex tasks. In addition to their remarkable performance in reasoning tasks, multi-agent systems have recently been explored for simulating human interaction. In this work, SpeechAgents \cite{DBLP:journals/corr/abs-2401-03945} employs a multi-modal LLM as the central control unit for individual agents, utilizing multi-modal signals as the medium for message exchange. Ultimately, the system is capable of simulating human dialogues with accurate content, authentic rhythm, and rich emotional expressions.

\section{Proposed Framework}

We propose a hybrid agent-based framework for multi-party dialogue synthesis, consisting of three distinct agents: \textbf{Script Writer}, responsible for generating and refining dialogue scripts based on character descriptions and feedback to ensure they are well-suited for conversational scenarios; \textbf{Speech Synthesizer}, which converts the scripts into natural and emotionally expressive speech by fully utilizing paralinguistic cues embedded in the scripts; \textbf{Dialogue Critic}, which reviews the synthesized dialogue and provides feedback in the form of textual instructions, subsequently feeds back to the Script Writer. Through the collaborative interaction of these three agents, the framework iteratively enhances the quality of the synthesized dialogues. As illustrated in Figure~\ref{fig:framework}, the black lines represent the initial dialogue generation process, while the red loop highlights the iterative optimization process.

\subsection{Character Pool Construction}

Previous dialogue datasets primarily involve conversations between two fixed speakers . To enhance the diversity of the generated dialogues, we collect a set of distinctive audios $Pool_a = \{a_i, i=1,2,...,N\}$, where $N$ is the number of the audios. Based on each audio $a_i$, we manually craft a character $c_i$ with profile $p_i$ detailing her/his age, gender, personality, and linguistic habits. These profiles collectively form the $Pool_p=\{p_i, i=1,2,...,N\}$. We further assign social relationships among different characters, including \textit{kinship}, \textit{friendship}, and \textit{colleague relationship}. As a result, we obtain a character pool $Pool_c$ = \{$c_1$, $c_2$, ... $c_n$ \}, where each $c_i$ includes $a_i$ and $p_i$. This role-playing approach is more conducive to generating more coherent and realistic dialogue scripts.
\subsection{Hybrid Agents}
\textbf{Script Writer.} The Script Writer agent $A_{w}$ typically possess excellent text generation and instruction following capabilities. Given the profile pool $Pool_p$, we prompt $A_{w}$ to generate a multi-party multi-turn dialogue script, 
\begin{equation*}
S = A_{w}(p_{A_{w}}, Pool_p)
\end{equation*}
where script \( S = \{c_j : u_j, \, j = 1, 2, \dots, M, \, c_j \in Pool_c\} \), \( u_j \) is the \( j \)-th utterance spoken by character \( c_j \), \( M \) denotes the number of utterances in the script, and the prompt \( p_{A_{w}} \) is as follows: % $||$ denotes concatenation, 
\begin{center}
\vspace{-1.2em}
\setlength{\fboxsep}{3pt} 
\fcolorbox{black}{gray!10}{\parbox{1\linewidth}{
\footnotesize Given character pool \{$Pool_c$\}, generate a conversation script meeting the below requirements:\\
1. \textbf{Character Interactions}: Characters should express distinct viewpoints or goals, with the ability to interrupt, question, or support others.\\
2. \textbf{Natural Flow}:
Dialogue should resemble natural communication, avoiding mechanical or overly deliberate language.
\\
3. \textbf{Topics and Content}: 
Dialogues should cover diverse, realistic topics, aligning with the character's setting and thematic background.\\
4. \textbf{Emotional Dynamics}: Incorporate emotional shifts to enhance expressiveness, using emotions such as happiness, anger, and sadness.
}}
\vspace{-0.5em}
\end{center}
\textbf{Speech Synthesizer.} Upon receiving the dialogue script generated by $A_{w}$, the Speech Synthesizer agent $A_{s}$ synthesizes high-fidelity speech dialogue with the guidance of character audios $Pool_a$,
\begin{equation*}
D = A_{s}(S, Pool_a)
\end{equation*}
\textbf{Dialogue Critic.} This critic agent $A_{c}$ is capable of comprehensively understanding the dialogues synthesized by $A_{s}$ and giving the feedback according to the prompt $p_{A_{c}}$ across multiple dimensions, with particular attention to the naturalness of the script content and emotional expression. 
\begin{equation*}
F = A_{c}(D, p_{A_{c}})
\end{equation*}
The prompt $p_{A_{c}}$ is as follows:
\begin{center}
\vspace{-3em}
\setlength{\fboxsep}{3pt} 
\fcolorbox{black}{gray!10}{\parbox{\linewidth}{
\footnotesize Please listen to this conversation sentence by sentence, evaluate each sentence based on the following criteria, and provide improvement suggestions:\\
1. \textbf{Naturalness:} Assess the smoothness and intonation, ensuring it matches the context with seamless transitions.\\
2. \textbf{Clarity and Emotiveness:} Ensure the sentence is clear, with no vague terms, and that emotional expression in the dialogue is sufficient and appropriate.
}}
\vspace{-0.5em}
\end{center}

%It then provides feedback $feedback\_{refine}$ in the form of textual instructions for subsequent iterative optimization of the framework.

\subsection{Collaborative Framework}

At time \( t=0 \), the Script Writer agent \( A_{w} \) generates the initial dialogue script \( S_0 \). The Speech Synthesize agent \( A_{s} \) then produces the corresponding dialogue \( D_0 \) based on \( S_0 \). After thoroughly analyzing \( D_0 \), the Dialogue Critic agent \( A_{c} \) provides feedback \( F_0 \). During the \( t \)-th iteration of the optimization process, where \( t \in [1, 2, \dots, T] \) and \( T \) is the total number of iterations, the Script Writer agent \( A_{w} \) receives feedback \( F_{t-1} \) and integrates it with the script \( S_{t-1} \) from the previous iteration to generate the revised dialogue script \( S_{t} \),
\begin{equation*}
S_{t} = A_{w}(p_{A_{w}}, Pool_p, S_{t-1}, F_{t-1})
\end{equation*}
Compared to $S_{t-1}$, $A_{w}$ modifies the dialogue content based on $F_{t-1}$, and add suitable paralinguistic tokens and emotional labels to each utterance, enhancing the overall coherence and realism of the dialogue. Subsequently, the $A_{s}$ produces a new dialogue $D_{t}$ based on the revised script $S_{t}$, and $A_c$ generates the feedback $F_t$. After \( T \) iterations, the final synthetic dialogue data \( \{S_T, D_T\} \) is produced.

\subsection{Framework Analysis}

\begin{table*}[t!]
    \centering
\renewcommand\tabcolsep{3.6pt}
\caption{Results of the speech evaluation.} 
\label{tab:speech_evaluation}
\resizebox{0.8\textwidth}{!}{
\begin{tabular}{lccccccc}
\toprule
\multirow{4}*{Model} &  \multicolumn{7}{c}{Speech} \\\cmidrule{2-8}
& \multicolumn{3}{c}{Human Evaluation } & \multicolumn{4}{c}{Automated Evaluation} \\
& MOS$\uparrow$& EMOS$\uparrow$ & TMOS$\uparrow$ & UTMOS(EN)$\uparrow$&UTMOS(CN)$\uparrow$ & WER(EN)$\downarrow$ & CER(CN)$\downarrow$ \\
\midrule
Writer + Synthesizer & 3.63\scriptsize{$\pm$0.060} & 3.64\scriptsize{$\pm$0.155}& 3.59\scriptsize{$\pm$0.034} & 4.186&2.962&4.22&5.19\\
Writer (self-refine) + Synthesizer & 3.63\scriptsize{$\pm$0.078}& 3.62\scriptsize{$\pm$0.061}&3.60\scriptsize{$\pm$0.064} & 4.224&3.043 &4.13&5.28 \\
Writer + Synthesizer + Critic (1 loop) &3.71\scriptsize{$\pm$0.063} &3.79\scriptsize{$\pm$0.041}& 3.71\scriptsize{$\pm$0.028} & 4.302& 3.138&4.14 &5.23 \\
Writer + Synthesizer + Critic (2 loop) &3.75\scriptsize{$\pm$0.071}&\textbf{3.96}\scriptsize{$\pm$0.043}& \textbf{3.78}\scriptsize{$\pm$0.037} &\textbf{4.316} & \textbf{3.318}&\textbf{4.07}&\textbf{5.16} \\
Writer + Synthesizer + Critic (3 loop) &\textbf{3.78}\scriptsize{$\pm$0.075} &3.91\scriptsize{$\pm$0.059}& 3.76\scriptsize{$\pm$0.142} &4.162 &3.064&4.34&5.36\\
\bottomrule
\end{tabular}
}
\end{table*}

\begin{table}[t]
\centering
\renewcommand\tabcolsep{3.6pt}
\caption{Results of the script evaluation.} % 表格标题
\label{tab:script_evaluation}
\resizebox{0.4\textwidth}{!}{
\begin{tabular}{lcc}
\toprule
Model & Naturalness$\uparrow$ & Emotiveness$\uparrow$ \\
\midrule
Writer + Synthesizer &4.32\scriptsize{$\pm$0.127}
&2.69\scriptsize{$\pm$0.677} \\
Writer (self-refine) + Synthesizer &4.45\scriptsize{$\pm$0.115}  &3.42\scriptsize{$\pm$0.552}\\
Writer + Synthesizer + Critic (1 loop) & 4.49\scriptsize{$\pm$0.102}& 3.67\scriptsize{$\pm$0.527}\\
Writer + Synthesizer + Critic (2 loop) & \textbf{4.59}\scriptsize{$\pm$0.088}&\textbf{3.96}\scriptsize{$\pm$0.320}\\
Writer + Synthesizer + Critic (3 loop) & 4.54\scriptsize{$\pm$0.070}&3.66\scriptsize{$\pm$0.344}\\

\bottomrule
\end{tabular}
}
\end{table}

1) Scripts are mostly manually written by professionals in previous dialogue datasets, which incurs high costs. Our framework leverages large language models' contextual understanding and instruction-based optimization to efficiently and sustainably generate natural and fluent dialogue scripts.

2) Our framework connects script generation and speech synthesis by introducing a dialogue critic agent, forming an iterative cycle for optimizing synthesized speech. This approach enables continuous improvement of both the dialogue scripts and the synthesized audio.

3) Our framework allows for flexible adaptation of different models for each agent, which could potentially improve the framework's synthesis capabilities.

\section{Framework Evaluation}
\subsection{Evaluation Details}

\noindent \textbf{Baselines.} To validate the effectiveness of DialogueAgents, we adopt five different variants of agent collaboration and synthesize 30 dialogues to evaluate performance. Initially, all variants begin with the same dialogue scripts generated solely by the script writer agent, which allows us to assess how incremental refinements impact the quality of the synthesized dialogues. The \textit{Writer + Synthesizer} variant represents dialogues synthesized directly from these original scripts without any refinement. In the \textit{Writer (self-refine) + Synthesizer} variant, the Script Writer agent performs a self-iteration to enhance the naturalness and emotional content of the dialogue script before synthesis. For the \textit{Writer + Synthesizer + Critic (1 loop)}, \textit{Writer + Synthesizer + Critic (2 loops)}, and \textit{Writer + Synthesizer + Critic (3 loops)} variants, the Script Writer agent receives refinement instructions from the Dialogue Critic agent, who evaluates the synthesized speech for understanding and quality. The Script Writer then iteratively optimizes the dialogue script based on this feedback. The term loop indicates different numbers of iterative cycles.  

\noindent \textbf{Character Pool Construction.} We collect 30 distinctive characters from the open-source audio data WenetSpeech4TTS \cite{ma2024wenetspeech4tts} and Common Voice \cite{ardila2019common}, and carefully crafted descriptive information for each character, covering aspects including age, personality, speaking style, and the social relationships between different characters. More details about the characters and descriptions can be referred to in our open-source project.
 
\noindent \textbf{Selection for Framework Agents.} We %considered both text generation capabilities and instruction optimization, 
adopt GPT-4o as the script writer agent, leveraging its advanced text generation capabilities and strong adherence to instructions. For dialogue generation, we utilize CosyVoice\footnote{\url{https://github.com/FunAudioLLM/CosyVoice}}, an advanced zero-shot text-to-speech synthesizer capable of instructed generation and fine-grained control over emotional and paralinguistic features. Qwen2-Audio serves as the dialogue critic agent, offering robust capabilities for processing various audio signal inputs and performing comprehensive audio analyses. %The number of iterations is set to $1$ by default.
%\input{tab/tab2}

%\subsection{Evaluation Metrics}

\noindent \textbf{Metrics.} For speech, we use the mainstream Mean Opinion Score (MOS) to assess the clarity and fluency of the dialogues. Moreover, we introduce two dialogue-level speech evaluation metrics: Emotional MOS (EMOS) and Turn-taking MOS (TMOS). EMOS evaluates the appropriateness and consistency of emotional and intonational changes within the dialogue, while TMOS assesses the naturalness and smoothness of speaker transitions. During testing, we invite 20 volunteers fluent in both English and Chinese to rate dialogue samples on a scale of 1 to 5. To ensure objective evaluation, we employ UTMOS \cite{DBLP:conf/interspeech/SaekiXNKTS22}, a MOS prediction system that leverages contrastive learning and phoneme encoding for automated and precise assessment of dialogue quality. Furthermore, we rigorously evaluate the synthesized dialogues through transcription-based analyses of Word Error Rate (WER) and Character Error Rate (CER). For text scripts, we employ an LLM (GPT-4o) as evaluator to assess naturalness and emotiveness, which is prompted to determine whether the scripts are natural and whether the emotional content is accurate and appropriate. 

\begin{figure}[t]  % 通栏的带*号
    \centering
    \includegraphics[width=0.24\textwidth]{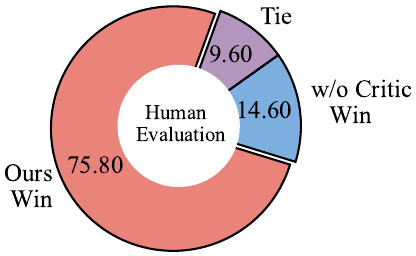}
    % \vspace{-0.5cm}
    \caption{Ablation of critic agent.}
    \label{fig:pie}
    % \vspace{-0.3cm}
\end{figure}

\subsection{Experimental Analysis}

\begin{table*}[t!]
	\caption{An example dialogue script. Refined scripts enhance the original versions by incorporating suitable paralinguistic and emotional tokens and refining the conversational content, resulting in more natural and expressive dialogue.}
	\centering
 \scriptsize
	%\scalebox{0.8}{
\begin{tabularx}{\textwidth}{X|X}
		\toprule
		script before refinement & 	script after refinement\\\midrule
\textit{James Carter:} You know, financial literacy is crucial for young adults today. Don’t you agree, Mark? & \textit{James Carter:}You know, financial literacy is \textbf{\textless strong\textgreater} so \textbf{\textless/strong\textgreater}crucial for young adults today. Wouldn't you agree, Mark? \textbf{[Engaging]}\\
\textit{Mark Williams:} Absolutely, James. Understanding budgeting and saving early can set them up for success in the future.
 & \textit{Mark Williams:} Absolutely, James! \textbf{[breath]} Understanding budgeting and saving early can really set them up for success in the future. It's so important! \textbf{[Agreeable]}\\
\textit{James Carter:} Right! But how do we make these concepts engaging for them?
 & \textit{James Carter}: Right! \textbf{[breath]} But how do we make these concepts \textbf{\textless strong\textgreater} engaging \textbf{\textless/strong\textgreater} for them? Maybe we could use practical examples and interactive tools. \textbf{[Curious]}\\
\textit{Mark Williams: } We could use practical examples and interactive tools. Let them manage a mock portfolio, perhaps.
 & \textit{Mark Williams: } We could let them manage a mock portfolio, perhaps. [breath] Real-life simulations could boost their interest and understanding. \textbf{[Innovative]}\\
\textit{James Carter:} That’s a solid idea, Mark. Real-life simulations can boost their interest and understanding.&\textit{James Carter:} That's a solid idea, Mark. Those hands-on experiences can make all the difference! \textbf{[Encouraged]}.\\
\bottomrule
\end{tabularx}
%}
\label{tab:Case}
\end{table*}

\begin{table}[t]
\caption{Statistics of MultiTalk Dataset.}
\begin{center}
\resizebox{0.4\textwidth}{!}{
\label{table:Statistics}
\begin{tabular}{lcccc}
\toprule
\multirow{2}*{Feature} & \multicolumn{2}{c}{CN} & \multicolumn{2}{c}{EN}\\
&AVG&Total&AVG&Total\\\midrule
Utterances   & 4.81  & 1950 & 4.41 & 2487 \\
Tokens & 115.47 &57,737  &86.07 & 43,036  \\
Duration (sec)&28.17&14,085.73&36.71&18,355.47\\
Roles & 3.54 &15  & 3.06  &15\\
\bottomrule
\end{tabular}}
\end{center}
\end{table}

%As shown in Table~\ref{tab:speech_evaluation}, \textit{Writer + Synthesize} represents the synthesis of dialogue based on the original dialogue script generated solely by the Script Writer agent. \textit{Writer (self-refine) + Synthesize} indicates that the Script Writer agent performs a self-iteration to refine the dialogue script before Speech Synthesizer using it for dialogue synthesis. This process is limited to the textual level of the dialogue script. \textit{Writer + Synthesize + Critic} refers to the process in which the Script Writer agent, after receiving refine instructions from the Dialogue Critic agent based on the understanding and evaluation at the level of synthesized dialogue speech, optimizes the dialogue script.

We evaluate the effectiveness of different agents as well as the overall iterative optimization capability, from both speech (Table~\ref{tab:speech_evaluation}) and script (Table~\ref{tab:script_evaluation}) perspectives.

\noindent \textbf{Script writer agent understands and follows instructions effectively.} It can be observed that \textit{Writer (self-refine) + Synthesizer} outperforms \textit{Writer + Synthesizer} across both speech and script metrics, indicating the script writer's ability to effectively follow instructions and further improve the script. Moreover, introducing the critic agent leads to further enhancements in the quality of the synthesized data, proving that the script writer effectively follows the critic's reviews to revise the script.

\noindent \textbf{Dialogue critic agent boosts the naturalness and emotiveness of script and speech.} As shown in the third row of Table~\ref{tab:speech_evaluation} and Table~\ref{tab:script_evaluation}, we find that incorporating feedback from the dialogue critic enhances the naturalness and emotiveness of the scripts, as well as the MOSes of speech. In addition, we pair the data synthesized by the \textit{Writer + Synthesize + Critic (1-hop)} and \textit{Writer + Synthesize} for human evaluation to determine which one is overall better. As shown in Figure~\ref{fig:pie}, the refined scripts outperform the originals in 75.8\% of cases, underscoring the importance of the critic’s role. %the results further prove the superior quality of scripts refined with the assistance of the dialogue critic agent, underscoring the importance of the critic’s role.

%To mitigate the potential instability introduced by LLM (GPT-4o) evaluations, we adopt a paired comparison approach. The original scripts generated by the Script Writer agent are paired with scripts via a full iteration of refinement within our synthesis framework. We then invite the same volunteers from the MOS evaluation to decide which script aligns better with real-world dialogue scenarios. We believe this paired comparison offers greater reliability. As shown in Figure~\ref{fig:pie}, the refined scripts outperform the originals in 75.8\% of cases, demonstrating not only that the Dialogue Critic agent enhances the naturalness and emotiveness of both scripts and speech but also that the Script Writer effectively follows the refinement instructions from the Dialogue Critic.

\noindent \textbf{The iterative nature of our framework further enhances the synthesis quality of dialogues.} When the three agents perform  two iterations  of collaborative refinement, the overall quality of the synthesized dialogues reaches optimal performance across most metrics. However, when the number of iterations increases to three, the dialogue quality begins to slightly decline due to noise and interference introduced by over-optimization. These findings suggest an optimal number of iterations exists that maximizes the quality of synthesized dialogues without incurring negative effects from over-processing.

%When the number of iterations reaches two, the overall quality of the synthesized dialogue achieves optimal performance across most metrics, whici demonstrates that our synthesis framework can significantly enhance dialogue quality, particularly in terms of emotional expression and dialogue cohesion, when refined within a reasonable scope. However, when the iteration count increases to three, a decline in dialogue quality is observed. This is likely due to excessive iterations leading to over-optimization, which introduces noise and interference. Despite this, the performance remains superior compared to the unrefined results.

%As the iteration number increases, the overall quality of the synthesized dataset consistently improves, demonstrating the superiority of the iterative generation process within the framework.

%\textbf{A reasonable increase in the number of iterations enhances the dialogue synthesis quality of the framework.} When the number of iterations reaches two, the overall quality of the synthesized dialogue achieves optimal performance across all metrics, except for MOS. This demonstrates that our synthesis framework can significantly enhance dialogue quality, particularly in terms of emotional expression and dialogue cohesion, when refined within a reasonable scope. However, when the iteration count increases to three, a decline in dialogue quality is observed. This is likely due to excessive iterations leading to over-optimization, which introduces noise and interference. Despite this, the performance remains superior compared to the unrefined results.

\section{Dataset Statistics and Analysis}

Building on the above analysis demonstrating the superiority of the DialogueAgents framework (comprising a writer, generator, and critic with two iterations), we generate MultiTalk—a multi-party, multi-turn dialogue dataset. This section presents detailed statistical analysis and evaluation of MultiTalk.

\subsection{Statistical Analysis} 

\begin{figure}[t]
    \centering
    \includegraphics[width=0.82\linewidth]{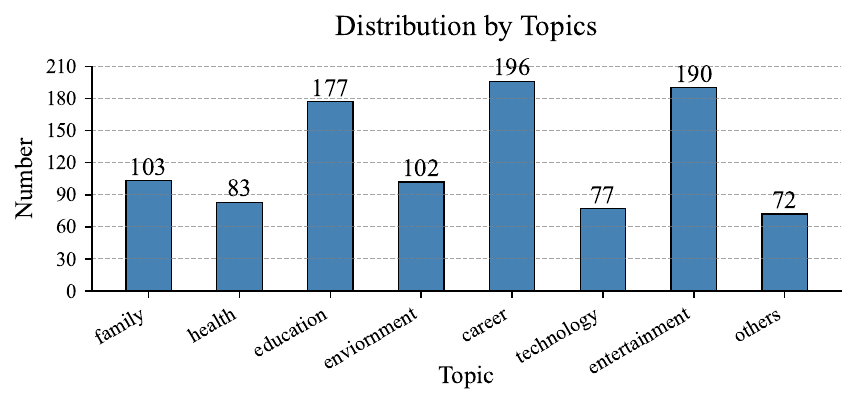}
    \caption{Distribution of the dataset by topics.}
    \label{fig:topics}
    % \vspace{-0.3cm}
\end{figure}

Table~\ref{table:Statistics} reports the key statistics of the MultiTalk, including both Chinese (CN) and English (EN) dialogues. The dataset consists of $4,437$ utterances, with an average of $4.44$ utterances per dialogue. The total duration of the dialogues is $32,441.2$ seconds, with the Chinese dialogues lasting $14,085.73$ seconds and the English dialogues $18,355.47$ seconds. Each Chinese and English dialogue involves an average of \textbf{$3.54$} and \textbf{$3.06$} character, respectively, with a total of $30$ distinct character.
Additionally, the dialogues in our dataset encompass diverse topics compared to previous datasets, with specific statistics shown in Figure~\ref{fig:topics}.
\subsection{Example Dialogue}

%[before]Here we present an example of a dialogue script before and after refinement as showed in Table~\ref{tab:Case}. After receiving the initial multi-party multi-turn dialogue script, the framework improves the dialogue text by incorporating suitable paralinguistic interpolation tokens and emotional labels, making it more natural and fluid.

In this section, we present an example to demonstrate how our proposed dialogue synthesis framework effectively refines the multi-party, multi-turn dialogue generation process. As illustrated in Table~\ref{tab:Case}, the initial dialogue script produced by the Script Writer contains only the conversational content between different speakers, without incorporating paralanguage elements such as pauses, emphasis, or emotional tone. This script is then fed into the Speech Synthesizer to generate the dialogue, which is subsequently evaluated at the dialogue level by the Dialogue Critic. Based on the evaluation, optimization recommendations are provided for each speaker's utterances and communicated in text form to the Script Writer, who revises the dialogue script accordingly.

The refined dialogue text exhibits significant improvements in three key dimensions, making it more closely aligned with natural conversational scenarios. First, the dialogue text is enhanced by substituting certain words and sentence structures with expressions that better fit the conversational context. Second, with respect to speech tone, suitable paralinguistic tokens are strategically introduced, including ``\textless strong\textgreater'' and ``[breath]'', improving the rhythm and flow of the dialogue. Finally, the integration of emotional tokens like ``[Agreeable]'' and ``[Curious]'' facilitates smoother emotional transitions between speakers, ensuring that the emotional tone of the dialogue accurately reflects the content. Following this refinement process, the synthesized dialogue not only becomes more representative of real-world conversational settings in terms of content, but also maintains a higher level of emotional coherence.

\subsection{Dataset Evaluation}
\begin{table}[t]
\caption{Results of the dataset evaluation.}
\begin{center}
\label{tab1}
\renewcommand\tabcolsep{3pt}
\resizebox{0.45\textwidth}{!}{
\begin{tabular}{lcccc}
\toprule
\multicolumn{1}{l}{Model} & VCTK                    & LJSpeech               & DailyTalk & Ours \\
\midrule
Ground Truth& 3.79\scriptsize{$\pm$0.042} & 3.82\scriptsize{$\pm$0.025} & 3.78\scriptsize{$\pm$0.038} & 3.76\scriptsize{$\pm$0.046}    \\
Tacotron2\cite{DBLP:conf/icassp/ShenPWSJYCZWRSA18} & 3.65\scriptsize{$\pm$0.090} & 3.71\scriptsize{$\pm$0.061} & 3.69\scriptsize{$\pm$0.036} & 3.67\scriptsize{$\pm$0.053}    \\
FastSpeech2\cite{DBLP:conf/iclr/0006H0QZZL21} & 3.72\scriptsize{$\pm$0.066} & 3.74\scriptsize{$\pm$0.026} & 3.70\scriptsize{$\pm$0.019} &
3.69\scriptsize{$\pm$0.038}    \\
DailyModel\cite{DBLP:conf/icassp/LeePK23}  & -- &  --&
3.67\scriptsize{$\pm$0.029} & 3.66\scriptsize{$\pm$0.045} \\   
\bottomrule
\end{tabular}%
}
\label{tab:comparison}
\end{center}
\end{table}
To demonstrate the usability of our synthesized dataset, MultiTalk, we compare it with existing speech datasets by evaluating their performance in training various models. Following the experimental setting of DailyTalk, We select two text-to-speech datasets, VCTK\cite{DBLP:conf/ococosda/VeauxYK13} and LJSpeech\footnote{\url{https://keithito.com/LJ-Speech-Dataset/}}, along with the dialogue dataset DailyTalk. Three representative models are employed: the auto-regressive model Tacotron2 \cite{DBLP:conf/icassp/ShenPWSJYCZWRSA18}, the non-autoregressive model FastSpeech2 \cite{DBLP:conf/iclr/0006H0QZZL21}, and the dialogue context-aware model DailyTalk-Model. All models are trained on a Tesla A40 GPU with a batch size of $16$, using the original hyperparameter configurations. Tacotron2 and FastSpeech2 are trained for 900K steps, while the DailyTalk-Model is trained on 850 dialogues and validated on 75 dialogues until convergence. As shown in Table~\ref{tab:comparison}, the MOS of the synthesized speech is comparable when models are trained on our dataset versus other datasets. Notably, compared to the dialogue dataset DailyTalk, MultiTalk achieves a similar MOS value despite having less than a quarter of the data volume. This further demonstrates the high quality of the dialogue data synthesized by our framework.

\section{CONCLUSION}
Agent technology has profoundly transformed traditional methods of conversational speech synthesis. In this paper, we present a novel Hybrid Agent-Based Speech Synthesis Framework for Multi-Party Dialogue, which automates the entire process from dialogue script creation to conversational speech synthesis and iterative refinement through the collaboration of three specialized agents. Furthermore, using this framework, we generate MultiTalk, a multi-party multi-turn dialogue dataset rich in emotional expressiveness and diverse character roles. Experiments based on comprehensive metrics validate the effectiveness of the framework and the integrity of our dataset. We hope that our work will benefit the research field of speech dialogue systems by enabling researchers to synthesize customized, high-quality dialogue data and further enhance the performance of speech synthesis models.

\section*{Acknowledgment}

We thank the anonymous reviewers for their constructive feedback. This work was supported in part by the National Key Research and Development Program of China under Grant 2022YFF0902701; in part by the National Natural Science Foundation of China under Grants U21A20468, 62372058, U22A2026.
Xiang Li was supported by the BUPT Excellent Ph.D. Students Foundation under Grant CX2023224.
W.\ Wang was supported by Guangdong Provincial Key Lab of Integrated Communication, Sensing and Computation for Ubiquitous Internet of Things (No.2023B1212010007, SL2023A03J00934), Guangzhou Municipal Science and Technology Project (No.\ 2023A03J0003, 2023A03J0013 and 2024A03J0621).

\bibliographystyle{IEEEbib}
\bibliography{icme2025references}
\begin{comment}
\newpage
\appendix
\section{}

    \begin{multicols}{2} % 开始双栏布局，只适用于本代码段部分
\begin{center}
\fcolorbox{black}{gray!10}{\parbox{\linewidth}{
\small 
Character Pool \\
\textbf{For English Speakers:}\\
\begin{description}
  \item[Name:] James Carter
  \item[Relation:] James and Mark are colleagues at the bank; Megan is their friend from college.
  \item[Role:] A middle-aged man who works in a bank.
  \item[Speaking Description:] Speaks fast, often uses ‘you know’ as a filler word. He tends to emphasize key points by raising his voice slightly.
  \item[WAV File:] path1
\end{description}
2. \textbf{Clarity and Emotiveness:} Ensure the sentence is clear, with no vague terms, and that emotional expression in the dialogue is sufficient and appropriate.
}}
\end{center}
\end{multicols} % 结束双栏布局

\begin{table*}[t!]
	\caption{An example dialogue script.}
	\centering
 \scriptsize
	%\scalebox{0.8}{
\begin{tabularx}{\textwidth}{lXXX}
		\toprule
		name & 	relation & role & speaking\_describe\\\midrule
JamesCarter&James and Mark are colleagues at the bank; Megan is their friend from college. & A middle-aged man who works in a bank. & Speaks fast, often uses ‘you know’ as a filler word. He tends to emphasize key points by raising his voice slightly.\\
\bottomrule
\end{tabularx}
%}
\label{tab:Case}
\end{table*}

\end{comment}

\end{document}